\documentclass{article} 

\usepackage{iclr2027_conference,times}

\usepackage[utf8]{inputenc} 
\usepackage[T1]{fontenc}    
\usepackage{hyperref}       
\usepackage{url}            
\usepackage{booktabs}       
\usepackage{amsfonts}       
\usepackage{nicefrac}       
\usepackage{microtype}      
\usepackage{xcolor}         
\usepackage{pifont}

\usepackage{amsmath}

\usepackage[linesnumbered,ruled,vlined]{algorithm2e}

\definecolor{mydarkblue}{RGB}{0,0,160}

\SetCommentSty{mycommfont}

\usepackage{graphicx}
\usepackage{multirow}
\usepackage{adjustbox}
\usepackage{colortbl}
\usepackage{tabularx}
\usepackage{arydshln}

\usepackage{wrapfig}
\usepackage{float}
\usepackage{caption}

\usepackage{xspace} 

\newcommand{\model}{S$^2$COPE\xspace}

\title{Can Unlabeled Data Teach Foundation Models Interpretable Concepts?}
\title{S$^2$COPE: Self-Supervised Concept Discovery via Preference Learning}

\author{%
  Shilong Xiang \quad Zirui Zhang \quad Chengzhi Mao \\
  Rutgers University \\
  \texttt{\{shilong.xiang,zirui.zhang,chengzhi.mao\}@rutgers.edu}\\
  \vspace{2mm}
  \href{https://shilongxiang.github.io/S2COPE/}{\texttt{https://shilongxiang.github.io/S2COPE/}}
}

\iclrpreprintcopy 

\begin{document}

\maketitle


\begin{abstract}
Self-supervised learning can learn useful representations from unlabeled images. Yet these representations remain largely opaque, making it difficult to understand what visual concepts they encode. Can self-supervision instead produce useful visual concepts that are directly understandable to people? We introduce \model, a framework that represents images as sets of concepts expressed in natural language. A vision-language model proposes candidate concepts, and a self-supervised objective evaluates how well each concept describes the source image while distinguishing it from other images. This feedback is used to construct preferences and train the model to propose better concepts. Across eight diverse visual datasets, our self-supervised concept representations transfer without adaptation and improve downstream recognition accuracy over prior interpretable approaches. In medical imaging, our discovered concepts reveal spurious visual cues and enable a more robust classifier. These results show that self-supervision can discover useful, human-readable visual representations from unlabeled images.
\end{abstract}


\section{Introduction}
\label{sec:intro}

Self-supervised learning can discover useful visual representations without annotations~\citep{simclr,moco,dino,mae}, but what it discovers remains difficult to interpret. In scientific domains, from cellular pathology to astrophysics, the goal is often not only to classify observations, but also to understand the visual patterns that distinguish them. In many such settings, the relevant concepts may not be known in advance; more fundamentally, the category taxonomy itself may not yet be defined when the data are first collected. Identifying which visual concepts matter, without first defining categories or annotating images, is therefore itself part of the discovery process~\citep{nature_ai,physical_concepts}. This motivates our question: \emph{how can models discover useful, human-readable visual concepts directly from unlabeled images?}




Prior work on interpretable visual representations often relies on predefined, annotated concepts~\citep{conceptbottleneck} or image category labels~\citep{label_free_cbm,mia}. Yet such supervision is not always available: concept annotations can be costly to obtain, and in many settings even the relevant category labels may not be known in advance. More recent approaches leverage embedding models~\citep{clip,zhai2023sigmoid,tschannen2025siglip2} or vision large language models (VLLMs)~\citep{qwen3,qwen3vl} to construct interpretable representations. These methods, however, typically treat the concept proposer as fixed: they use its descriptions for downstream prediction, but do not learn from the target images which concepts are most useful to propose~\citep{sachit,label_free_cbm,labo}. What is missing is a source of feedback that can teach the concept generator, directly from unlabeled images, which concepts are useful—without relying on predefined classes, annotated concepts, or human judgments.

\begin{figure*}[t]
\centering
\vspace{-5mm}
\includegraphics[width=0.98\textwidth, trim={0.8cm 0 0.7cm 0}, clip]{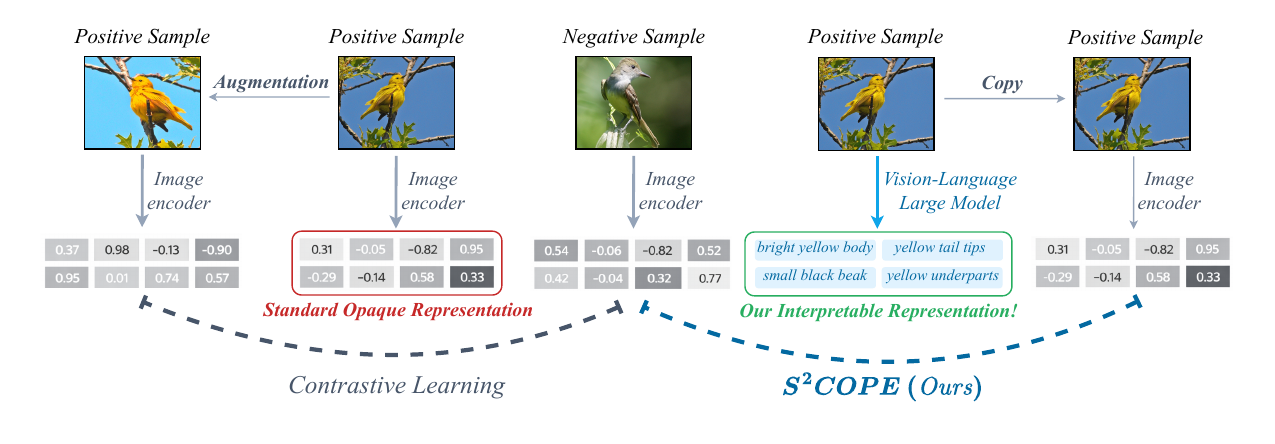}
\vspace{-6mm}
\caption{\textbf{Self-Supervised Visual Concept Discovery.} (Left) Standard self-supervised learning represents images as feature vectors, which are not directly understandable.
(Right) \model learns to propose human-readable visual concepts from unlabeled images.
A pretrained vision-language model samples candidate concepts, and a contrastive objective evaluates how well they describe the source image and distinguish it from other images.
These scores define preferences for training the concept generator, without class labels or concept annotations.
The learned concepts support downstream prediction and provide an explicit vocabulary for inspecting model behavior.
}
\vspace{-2mm}
\label{fig:teaser}
\end{figure*}

In this paper, we introduce \model, a framework that learns to generate interpretable concepts from unlabeled data (Figure~\ref{fig:teaser}). Our key insight is that the same self-supervised signal used to learn visual representations can also teach a model which language concepts are useful to generate. Starting from a pretrained vision-language model, we sample candidate concept sets and evaluate them with a self-supervised contrastive objective that measures how well a concept set describes its source image while distinguishing it from other images. Higher- and lower-scoring candidates are then paired as preferences and used to train the concept generator with Direct Preference Optimization (DPO)~\citep{dpo}. Rather than treating concept proposal as a fixed preprocessing step, \model adapts the model's proposals to the visual patterns present in the unlabeled training images. We call this process self-supervised because it requires neither target class names, image labels, concept annotations, nor human judgments.


We evaluate the learned concepts through two complementary applications: cross-domain recognition and shortcut mitigation. Across eight datasets spanning natural images~\citep{inaturalist,cub}, faces~\citep{karkkainen2021fairface}, medical imaging~\citep{ham10000,medmnist}, and physics~\citep{galaxy10,gravityspy}, our learned concepts outperform four concept-based baselines on every dataset, improving downstream top-1 accuracy by 7.37 percentage points on average over the best of these baselines per dataset. LLM-based and human evaluations further show that our method achieves the highest win rate in the quality of the discovered concepts. In ISIC dermoscopy~\citep{nauta2022shortcut,rieger2020interpretations}, our self-supervised concepts expose colored calibration patches associated with benign lesions. Filtering training images using these concepts reduces the patch-induced error rate on malignant images by 15 points. Together, these results suggest that concept discovery need not begin with category labels: a language model can propose candidate hypotheses, while unlabeled visual data provides the evidence needed to determine which concepts are useful.

\section{Related Work}
\label{sec:related_work}

\noindent\textbf{Self-Supervised Learning.}
Self-supervised learning shows that useful visual representations can be learned without image annotations~\citep{simclr,moco,dino,mae,byol,swav,vicreg,ibot,dinov2,deepcluster}.
The training signal comes from the images themselves, but what the model learns is expressed as feature vectors rather than concepts with explicit meanings.
Recent work brings related objectives into large vision-language models through reconstruction~\citep{ross}, jigsaw tasks~\citep{visual_jigsaw}, feature alignment~\citep{viral}, and auxiliary visual tasks~\citep{jarvis}.
These approaches improve visual learning without directly teaching the model which human-readable concepts to propose.
Our work applies self-supervision to concept discovery itself.
We evaluate candidate concept sets by how well they describe their source image and distinguish it from other images.

\noindent\textbf{Concept-Based Models.}
Concept bottleneck models make predictions through named visual properties, allowing users to inspect the concepts behind a decision~\citep{conceptbottleneck}.
The difficulty is choosing these concepts without first specifying what the model should look for.
Early models rely on human-defined vocabularies and concept annotations.
Later work reduces this requirement by using language models to propose descriptions~\citep{labo,label_free_cbm}, reusing pretrained concept representations~\citep{pcbm}, or linking concepts to their visual evidence~\citep{vlg_cbm_2024}.
However, removing concept annotations does not necessarily remove the need for known categories.
LaBo and LF-CBM use class names to guide concept generation~\citep{labo,label_free_cbm}, while PCBM can use class names to retrieve related concepts from a knowledge base~\citep{pcbm}.
These approaches therefore provide ways to build interpretable predictors for known categories, but do not directly address discovery when those categories have not been specified.

Another line of work extracts and describes concepts from the features of an existing model~\citep{sae_vlm_2025,mcbm_2025,ucbm_2025,clipdissect}.
U-F$^2$-CBM~\citep{uf2cbm} discovers concepts from frozen features without class labels.
M-CBM~\citep{de2026learning} uses a sparse autoencoder to extract features from a pretrained classifier, then a multimodal language model to name and annotate the resulting concepts.
Related to our shortcut analysis, B2T~\citep{kim2024discovering} extracts keywords from captions of misclassified images to identify spurious correlations; it uses class labels to identify these errors and keeps the captioner fixed.
We instead train the concept generator itself using feedback from unlabeled images, rather than only extract, name, or select concepts from a fixed model.
Neither concept generation nor selection in \model requires class names or image labels.

\noindent\textbf{Preference Learning.}
Vision-language pretraining and instruction tuning connect visual content with language~\citep{align,flamingo,blip2,instructblip,llava}.
However, a fluent description need not be accurate or capture the visual details that distinguish one image from another~\citep{hallucination_vlm,chair,hallusionbench,taxonomy_aware_vlm_eval}.
Preference learning offers a way to improve generated responses by training on comparisons between alternatives~\citep{vdpo_2025,chip_2025,realign_2025,spo_iccv2025,symdpo}.
A central question is how to obtain these comparisons without asking people or a stronger model to judge each output.
SeVa~\citep{seva} and SMPRO~\citep{smpro} construct visual preference data without external annotators to improve general-purpose multimodal responses.
Our goal is instead to learn which visual concepts to propose.
We score candidate concept sets by how well they describe their source image and distinguish it from other images.
These scores define preference pairs for training the concept generator through direct preference optimization.

\section{Methodology}
\label{sec:method}

\begin{figure*}[t]
\centering
\vspace{-5mm}
\includegraphics[width=1.0\textwidth, trim={0.8cm 0 0.8cm 0}, clip]{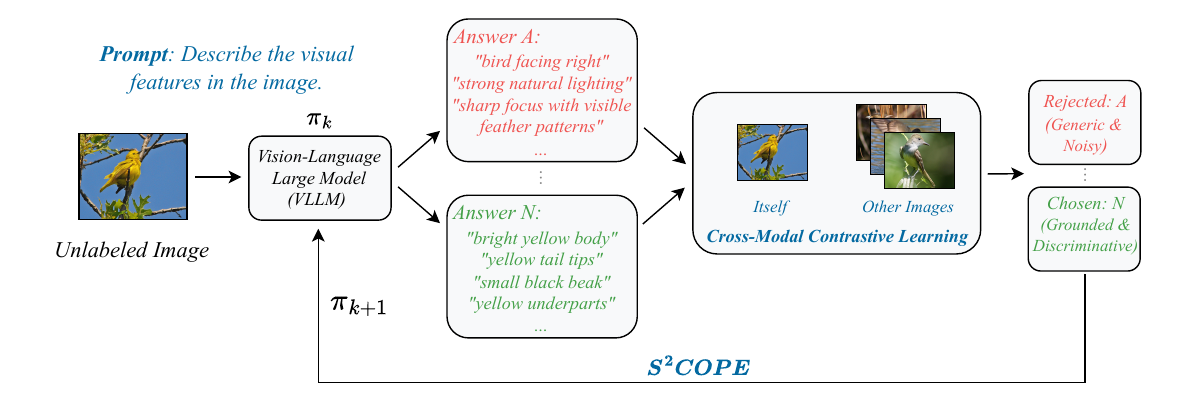}
\caption{
\textbf{Overview of the \model Discovery Loop.}
At iteration $k$, the generator $\pi_k$ samples several descriptions for each unlabeled image.
A frozen CLIP model scores each description by how well it matches the source image relative to other images.
We pair higher- and lower-scoring descriptions with the largest score gaps and train the generator on these preferences using DPO.
The updated generator $\pi_{k+1}$ then proposes candidates for the next iteration.
}
\label{fig:method_overview}
\end{figure*}

Standard representation learning maps images to feature vectors that are hard to interpret.
We instead seek a vocabulary of natural-language concepts as our representations.
\model learns this vocabulary through three steps: sampling descriptions from the generator, scoring them against the unlabeled images, and training the generator to increase the resulting preference scores (Figure~\ref{fig:method_overview}).
We start from pretrained VLLMs, but use no class names, image labels, or concept annotations from the training images during our concept discovery.

\subsection{Concept Proposal via Vision-Language Models}
\label{sec:candidate_generation}

For each image $x$, we prompt the current vision-language generator $\pi_k$, a VLLM (Qwen3-VL-8B-Instruct in our experiments), to describe its visual attributes and sample $N$ responses, $\mathcal{C}(x)=\{c_1,\ldots,c_N\}$.
We use high-temperature sampling to obtain varied candidates.
Each response may contain multiple concepts and is scored as a whole; we extract individual concepts later for downstream evaluation.

\subsection{Cross-Modal Contrastive Reward}
\label{sec:contrastive_reward}

For each image $x$, the generator produces candidate descriptions $\mathcal{C}(x)$.
We evaluate each candidate by how well it matches its source image relative to other images.
A separate, frozen CLIP model~\citep{clip,openclip} provides this comparison: its image encoder $f_{\mathrm{img}}$ encodes $x$, and its text encoder $f_{\mathrm{text}}$ encodes each description $c\in\mathcal{C}(x)$.
We normalize their outputs to obtain
\begin{equation}
    v_x =
    \frac{f_{\mathrm{img}}(x)}
    {\|f_{\mathrm{img}}(x)\|_2},
    \qquad
    t_c =
    \frac{f_{\mathrm{text}}(c)}
    {\|f_{\mathrm{text}}(c)\|_2}.
\end{equation}
Thus, all candidates for an image share the same image embedding $v_x$, while each candidate description $c$ has its own text embedding $t_c$.

For a batch $\mathcal{B}$ containing $x$, we encode every image in the same way and score each candidate as
\begin{equation}
    R(c,x)
    =
    \log
    \frac{\exp(t_c^\top v_x/\tau)}
    {\sum_{x'\in\mathcal{B}}\exp(t_c^\top v_{x'}/\tau)},
\end{equation}
where $\tau$ is a temperature.
The source image is the positive match, and the other batch images serve as negatives.
A description $c$ receives a high score when it matches its source image more strongly than the alternatives, rather than describing many images equally well. As described in the next subsection, this score is then used to train the generator to assign higher probability to such concepts. For the positive image corresponding to the query image being interpreted, we use the original image without augmentation. This differs from standard contrastive learning, where augmentations encode prior assumptions about which transformations the representation should ignore. In our setting, we do not assume in advance which visual variations the discovered concepts should be invariant to, and empirically find that using the original image works best. We study the effect of introducing additional augmentations in Table~\ref{tab:augmentation}.

\begin{figure*}[t]
\centering
\vspace{-5mm}
\includegraphics[
  trim={1.2cm 0.6cm 0.6cm 0.65cm}, clip,
  width=\textwidth,
  height=0.8\textheight,
  keepaspectratio
]
{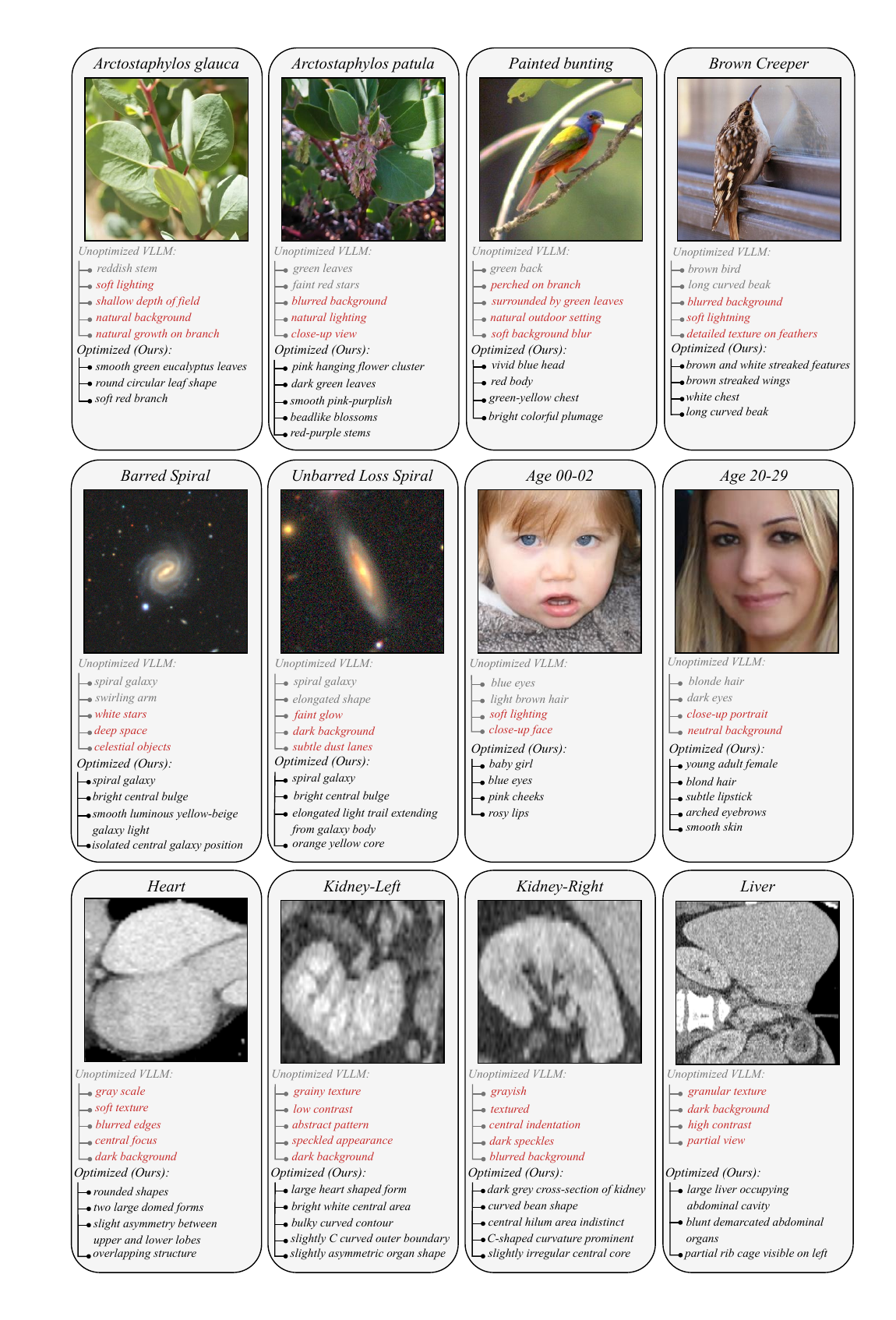}
\vspace{-4mm}
\caption{\textbf{Concepts Before and After Learning.}
For each image, we compare concepts from the pretrained generator (top list) and \model{} (bottom list).
\textcolor{red}{Red} marks concepts that are incorrect or uninformative for the category.
}
\vspace{-5mm}
\label{fig:qualitative_results}
\end{figure*}

\subsection{Self-Supervised Preference Optimization}
\label{sec:dpo_optimization}

Rather than only filtering proposals from a fixed generator~\citep{label_free_cbm}, we use the contrastive scores to improve what the VLLM generator proposes.
For each image, we select candidate pairs with the largest score gaps.
The higher-scoring description becomes $c_w$ and the lower-scoring description becomes $c_l$, forming a preference dataset $\mathcal{D}_k=\{(x,c_w,c_l)\}$.

Starting from $\pi_k$, we train only the generator $\pi_\theta$ on these pairs using Direct Preference Optimization (DPO)~\citep{dpo}:
\begin{equation}
\mathcal{L}_{\mathrm{DPO}} = -\mathbb{E}_{(x,c_w,c_l)\sim\mathcal{D}_k} \left[ \log \sigma \left( \beta \log \frac{\pi_\theta(c_w\mid x)}{\pi_{\mathrm{ref}}(c_w\mid x)} - \beta \log \frac{\pi_\theta(c_l\mid x)}{\pi_{\mathrm{ref}}(c_l\mid x)} \right) \right].
\end{equation}
Here, $\pi_{\mathrm{ref}}$ is a frozen reference model which is the generator $\pi_k$ from the previous iteration, $\sigma$ is the sigmoid function, and $\beta$ controls regularization toward the reference.
The objective teaches the generator to favor higher-scoring descriptions over lower-scoring ones.
We update the generator through the DPO loss, without differentiating through the sampled text or CLIP score.
Figure~\ref{fig:qualitative_results} compares concepts before and after training.

\paragraph{Iterative Optimization.} We repeat the above procedure for $K$ rounds. After each DPO update, the optimized vision-language model generates a new set of candidate concepts. These concepts are then evaluated using the same self-supervised objective, converted into preference pairs, and used for the next round of DPO training $\pi_{k+1}$. Repeating this cycle allows the concept generator to progressively adapt its proposals to the visual structure of the unlabeled training images.

\noindent\textbf{Evaluation.}
Following the linear evaluation protocol in self-supervised representation learning~\citep{simclr,moco,dino}, we test whether the learned concepts support classification without further adapting the representation.
After training, we freeze the model and generate descriptions for the training images and form a fixed concept bank.
Each image is represented by its CLIP similarities to the concepts in this bank.
We train a logistic regression classifier on these vectors and report top-1 accuracy on test images represented using the same bank.
Class labels are used to train a linear probe classifier to evaluate the quality of the interpretable representation, not to discover concepts or construct the bank.
Moreover, to assess the quality of the concepts, we report their win rates over the compared baselines, as judged by both a separate vision-language model and human evaluators. In addition, we show that removing training images flagged by the discovered bias concepts yields more robust classifiers on downstream tasks.

\section{Experiments}

\subsection{Experimental Setup}
\label{sec:experimental_setup}

\noindent\textbf{Datasets.} We optimize the concept generator on unlabeled \textbf{iNaturalist-mini} images~\citep{inaturalist}, following the split of~\citet{mia}, and evaluate frozen transfer across eight datasets in four domains. Except for the iNaturalist evaluation set, all datasets are unseen during training. \textbf{Nature}: \textbf{iNaturalist}~\citep{inaturalist,mia} and \textbf{CUB}~\citep{cub} (fine-grained biological categories). \textbf{Faces}: \textbf{FairFace}~\citep{karkkainen2021fairface} (nine age groups). \textbf{Medical}: \textbf{HAM10000}~\citep{ham10000} (skin lesions), \textbf{OrganCMNIST}, and \textbf{OrganMNIST3D}~\citep{medmnist} (CT scans). \textbf{Physics}: \textbf{Galaxy10}~\citep{galaxy10} (galaxy morphologies) and \textbf{Gravity Spy}~\citep{gravityspy} (gravitational-wave spectrograms).

\definecolor{gainblue}{RGB}{58, 118, 205}
\definecolor{rowgray}{gray}{0.94}

\begin{table}[t]
\centering
\small
\setlength{\tabcolsep}{3pt}
\vspace{-5mm}
\caption{\textbf{Top-1 Concept-Based Classification Accuracy (\%) Across Eight Datasets.} \model fine-tunes a VLLM on unlabeled iNaturalist-mini images, which is then transferred to other datasets without adaptation. ``Labels'' indicates class-label or class-name use in concept construction. \textcolor{gainblue}{Blue} values denote percentage-point gains over the best of the four concept-based baselines per dataset; bold marks the highest accuracy. \model outperforms these baselines on all eight datasets, with an average gain of 7.37 percentage points.}
\label{tab:main_results}

\begin{tabularx}{\textwidth}{l c : *{4}{>{\centering\arraybackslash}X}}
\toprule

\multirow{2}{*}{\textbf{Method}} & \multirow{2}{*}{\textbf{Labels}} & \multicolumn{2}{c}{\textbf{Nature}} & \multicolumn{2}{c}{\textbf{Physics}} \\
\cmidrule(lr){3-4} \cmidrule(lr){5-6}
& & \textbf{iNaturalist} & \textbf{CUB} & \textbf{Galaxy10} & \textbf{Gravity Spy} \\
\midrule
TextSpan & \ding{51} & 20.00 & 81.08 & 24.07 & 4.25 \\
DCLIP & \ding{51} & 23.46 & 82.14 & 11.75 & 2.17 \\
LF-CBM & \ding{51} & 73.85 & 64.55 & 53.44 & 83.08 \\
LaBo & \ding{51} & 42.69 & 77.63 & 53.55 & 39.50 \\
\midrule
VLLM (Unoptimized) & \ding{55} & 71.15 & 74.77 & 54.09 & 60.83 \\
\rowcolor{rowgray} \textbf{VLLM + \model (Ours)} & \ding{55} & \textbf{85.00} \textcolor{gainblue}{(+11.15)} & \textbf{85.11} \textcolor{gainblue}{(+2.97)} & \textbf{66.52} \textcolor{gainblue}{(+12.97)} & \textbf{83.55} \textcolor{gainblue}{(+0.47)} \\
\bottomrule
\end{tabularx}

\vspace{0.3em}

\begin{tabularx}{\textwidth}{l c : *{4}{>{\centering\arraybackslash}X}}
\toprule

\multirow{2}{*}{\textbf{Method}} & \multirow{2}{*}{\textbf{Labels}} & \multicolumn{3}{c}{\textbf{Medical}} & \textbf{Faces} \\
\cmidrule(lr){3-5} \cmidrule(lr){6-6}
& & \textbf{HAM10000} & \textbf{\scriptsize OrganCMNIST} & \textbf{\scriptsize OrganMNIST3D} & \textbf{FairFace} \\
\midrule
TextSpan & \ding{51} & 17.10 & 7.40 & 12.42 & 50.95 \\
DCLIP & \ding{51} & 33.16 & 10.41 & 14.29 & 46.16 \\
LF-CBM & \ding{51} & 70.98 & 78.85 & 65.22 & 53.47 \\
LaBo & \ding{51} & 63.73 & 54.10 & 22.36 & 39.99 \\
\midrule
VLLM (Unoptimized) & \ding{55} & 65.28 & 68.06 & 58.39 & 50.63 \\
\rowcolor{rowgray} \textbf{VLLM + \model (Ours)} & \ding{55} & \textbf{79.27} \textcolor{gainblue}{(+8.29)} & \textbf{91.43} \textcolor{gainblue}{(+12.58)} & \textbf{70.19} \textcolor{gainblue}{(+4.97)} & \textbf{59.01} \textcolor{gainblue}{(+5.54)} \\
\bottomrule
\end{tabularx}
\end{table}

\begin{table}[t]
\centering
\scriptsize
\caption{\textbf{Pairwise Preference Win Rate for \model Concepts (\%).}
For each opponent, we report the percentage of the 39 images for which the evaluator prefers \model's anonymized concept list over the opponent's.
GPT-5.6-Sol compares \model against five baselines, while human evaluators compare \model against the unoptimized VLLM, with results averaged across evaluators. Our evaluation confirms that \model is able to generate concepts that represent the image better.}
\vspace{-0.1cm}
\label{tab:vlm_judge}
\begin{tabular}{lcccccc}
\toprule
Evaluator & \multicolumn{5}{c}{GPT-5.6-Sol} & \textbf{Human} \\
\cmidrule(lr){2-6} \cmidrule(lr){7-7}
Opponent & Unoptimized VLLM & TextSpan & DCLIP & LF-CBM & LaBo & Unoptimized VLLM \\
\midrule
\model preference & 85.9 & 98.7 & 96.2 & 89.7 & 98.7 & \textbf{96.41} \\
\bottomrule
\end{tabular}
\end{table}

\begin{table}[t]
\centering
\small
\caption{\textbf{Concept-Guided Shortcut Mitigation on ISIC.} Colored calibration patches are spuriously associated with benign lesions. \model identifies affected training images; removing them and retraining the linear probe reduces the patch-induced error rate from 74.9\% to 59.9\%. \model thus enables targeted data curation to reduce shortcut reliance.}
\label{tab:spurious_results}
\begin{tabular}{lccc}
\toprule
\multirow{2}{*}{\textbf{Setting}} & \multicolumn{2}{c}{\textbf{Patch detection} (\%)} & \multirow{2}{*}{\textbf{Error rate} (\%) $\downarrow$} \\
\cmidrule(lr){2-3}
& \textbf{Precision} $\uparrow$ & \textbf{Recall} $\uparrow$ & \\
\midrule
VLLM (Unoptimized) & --- & --- & 74.9 \\
VLLM + \model & 95.5 & 83.4 & \textbf{59.9} \\
\bottomrule
\end{tabular}
\end{table}

\noindent\textbf{Baselines.} We compare against four supervised concept-based methods that also produce interpretable concepts. \textbf{TextSpan}~\citep{textspan} decomposes CLIP image representations into text-interpretable components via a spanning set of natural-language descriptions. \textbf{DCLIP}~\citep{sachit} performs zero-shot classification using LLM-generated per-class descriptions matched via CLIP. \textbf{LF-CBM}~\citep{label_free_cbm} prompts an LLM with class names to build a concept bank, then trains a linear classifier on CLIP concept activations. \textbf{LaBo}~\citep{labo} further applies submodular selection to retain a maximally discriminative concept subset. 


\noindent\textbf{Models.} For fair comparison, all methods, both baselines and ours, employ the same Qwen3-VL-8B-Instruct~\citep{qwen3,qwen3vl} for concept generation (except TextSpan, which uses its released general description set) and OpenCLIP ViT-H/14~\citep{openclip} for concept-image similarity scoring, isolating each method's concept discovery strategy and the contribution of \model from differences in model capacity.

\noindent\textbf{Implementation Details.} We use a batch size of 1,300 for the contrastive reward calculation. We optimize with batch size of 512. We iterate the learning by $K=10$ rounds. We use learning rate of 5e-6 for the Qwen3-VL vision tower and 1e-6 for the rest of Qwen3-VL parameters. Our training runs on one node with 8 RTX 6000 Blackwell Pro GPUs.

\begin{figure}[t]
    \vspace{-6pt}
    \centering
\hspace{-0.6cm}\includegraphics[width=0.73\textwidth, trim={1 0 0 0}, clip]{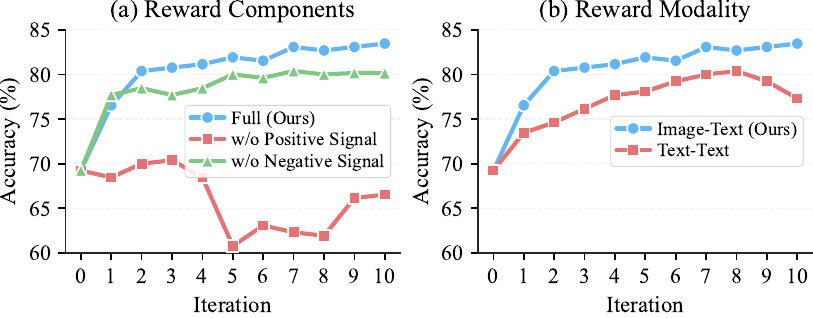}
    \caption{\textbf{Ablation Studies on Reward Formulation.}
    \textbf{(a) Reward Components:} Impact of isolating the positive and negative signals of the contrastive reward. Eliminating the positive signal causes a performance collapse, while removing the negative signal yields a suboptimal accuracy plateau.
    \textbf{(b) Reward Modality:} Comparison of cross-modal image-text grounding versus unimodal text-text consensus. Cross-modal grounding against physical image features achieves better performance than relying on unimodal textual consensus.}
    \label{fig:all_ablations}
    \vspace{1mm}
\end{figure}

\begin{table}[t]
\centering
\vspace{-2mm}
\begin{minipage}[c]{0.64\textwidth}
\centering
\small
\setlength{\tabcolsep}{4pt}
\renewcommand{\arraystretch}{1.08}
\begin{tabularx}{\linewidth}{@{}Xc@{}}
\toprule
\textbf{Augmentation} & \textbf{Accuracy (\%)} \\
\midrule
No augmentation & \textbf{83.46} \\
\quad + Horizontal flip, color jitter, Gaussian blur & 82.69 \\
\quad + Random crop & 82.31 \\
\quad + Grayscale, wider crop, stronger jitter and blur & 81.92 \\
\midrule
Random crop only & 80.77 \\
Random crop, horizontal flip only & 81.92 \\
Color jitter, grayscale, Gaussian blur only & 81.92 \\
\bottomrule
\end{tabularx}
\end{minipage}\hfill
\begin{minipage}[c]{0.30\textwidth}
\captionsetup{font=small,skip=0pt}
\caption{\textbf{Effect of Data Augmentation.} We vary the image augmentations used to compute the contrastive reward, adding them cumulatively (top) or evaluating them separately (bottom). Using the original images achieves the highest accuracy among the evaluated settings.}
\label{tab:augmentation}
\end{minipage}
\end{table}

\begin{figure}[t]
\centering
\begin{minipage}[c]{0.52\textwidth}
\includegraphics[width=\linewidth]{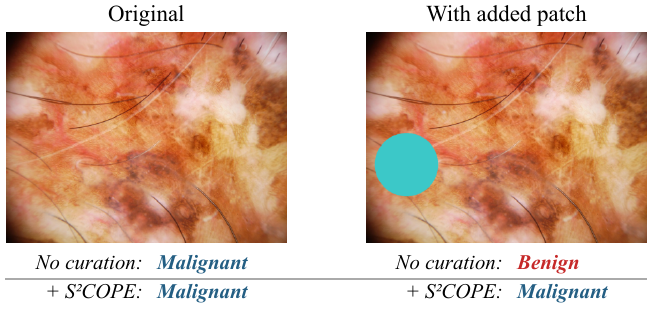}
\end{minipage}\hfill
\begin{minipage}[c]{0.45\textwidth}
\captionsetup{font=small,skip=0pt}
\vspace{-4mm}
\caption{\textbf{Concept-Guided Shortcut Mitigation.} Adding a patch causes this malignant image to be misclassified as benign. ``+ \model'' denotes concept-guided data curation and probe retraining, which preserve the correct prediction. \model enables targeted data curation that removes bias in the data.}
\vspace{-4mm}
\label{fig:spurious_example}
\end{minipage}
\end{figure}

\begin{table}[t]
\vspace{-3mm}
\begin{minipage}[c]{0.52\textwidth}
\centering
\scriptsize
\begin{tabular}{lcc}
\toprule
\multirow{2}{*}{\textbf{Base Model}} & \multicolumn{2}{c}{\textbf{Top-1 Accuracy (\%)}} \\
\cmidrule(lr){2-3}
& Unoptimized & + \model \\
\midrule
Qwen3-VL-2B-Instruct & 67.31 & \textbf{80.77} \\
Qwen3-VL-4B-Instruct & 69.62 & \textbf{82.31} \\
Qwen3-VL-8B-Instruct & 69.23 & \textbf{83.46} \\
\bottomrule
\end{tabular}
\end{minipage}
\hfill
\begin{minipage}[c]{0.45\textwidth}
\captionsetup{font=small,skip=0pt}
\caption{\textbf{Impact of Base Model Capacity.} Ablation over VLLM scale (2B, 4B, 8B). \model yields consistent gains (12.7--14.2 points) across all scales, with accuracy scaling monotonically with model size.}
\label{tab:base_model}
\end{minipage}
\end{table}

\begin{table}[t]
\centering
\scriptsize
\vspace{-4mm}
\caption{\textbf{Selection of the Negatives.} We compare three levels of hardness on the negative samples: Hard Neg\ (smallest gap), Random, and Easy Neg\ (largest gap). Each strategy selects 16 preference pairs per image. Easy negatives yield the best accuracy.}
\vspace{-0.1cm}
\label{tab:dpo_strategy}
\begin{adjustbox}{max width=\columnwidth}
\setlength{\tabcolsep}{3.5mm}
\begin{tabular}{lcccc}
\toprule
\multirow{2}{*}{\textbf{Strategy}} & \multirow{2}{*}{\textbf{Unoptimized VLLM}} & \multicolumn{3}{c}{\textbf{VLLM + \model}} \\
\cmidrule(lr){3-5}
& & \textbf{Hard Neg} & \textbf{Random} & \textbf{Easy Neg} \\
\midrule
\textbf{Top-1 Accuracy (\%)} & 69.23 & 76.92 & 81.92 & \textbf{83.46} \\
\bottomrule
\end{tabular}
\end{adjustbox}
\vspace{-4mm}
\end{table}

\subsection{Main Results}
\label{sec:main_results}

We evaluate two uses of the learned concepts: representing images for recognition and exposing shortcuts for targeted data curation. In both settings, the concept generator is trained only on unlabeled iNaturalist-mini images and transferred without adaptation.

\subsubsection{Cross-Domain Recognition and Preference Evaluation}

\paragraph{Probing Results.} Table~\ref{tab:main_results} reports concept-based classification accuracy across eight datasets. \model outperforms all four concept-based baselines on every dataset, with a mean gain of 7.37 percentage points over the best of these baselines on each dataset. Gains range from +0.47 points on Gravity Spy to +12.97 points on Galaxy10, with a similarly large gain on OrganCMNIST (+12.58 points), demonstrating transfer to domains absent from concept-generator training. \model also consistently improves over the unoptimized VLLM, supporting the benefit of self-supervised preference learning for concept discovery.

\paragraph{Pairwise Preference Evaluation.}
\label{sec:vlm_judge}
We evaluate concept quality through VLLM-based and human comparisons on a shared set of 39 images drawn from our evaluation domains (Table~\ref{tab:vlm_judge}). GPT-5.6-Sol prefers \model concepts over the unoptimized VLLM in 85.9\% of judgments and over each concept-based baseline in at least 89.7\%, with each pair evaluated in both A/B orders (Appendix~\ref{sec:judge_protocol}). Ten volunteers select the more informative description of the depicted object, preferring \model over the unoptimized VLLM at a mean rate of 96.41\%. These results support the improved descriptive quality of concepts learned through our self-supervised preference optimization.

\subsubsection{Discovering and Mitigating Spurious Correlations}
\label{sec:spurious_correlation}

In addition to predictive accuracy, which measures the capability of our discovered concepts, we further test whether the discovered concepts can expose and help mitigate data bias. We study ISIC dermoscopy, where colored calibration patches are associated with benign lesions~\citep{nauta2022shortcut,rieger2020interpretations}. On a 29,967-image ISIC dataset~\citep{codella2018skin}, our learned generator produces concepts such as ``orange circular marker,'' ``green circular patch,'' and ``round sticker-like texture'' without being supplied artifact names or patch annotations, which enables the inspection of the confounding factor.

In Table~\ref{tab:spurious_results}, ``Unoptimized'' denotes the pipeline before data curation, while ``+ \model'' denotes concept-guided filtering and probe retraining, where patch-group labels are used only to select which concept clusters to filter on, not to generate concepts; both use the same frozen VLLM and concept bank. Filtering identifies artifact-containing images with 95.5\% precision and 83.4\% recall. We measure the patch-induced error rate on malignant test images correctly classified before patch insertion. Retraining the probe on the filtered data reduces this rate from 74.9\% to 59.9\%, a 15.0-point reduction. Figure~\ref{fig:spurious_example} illustrates the resulting robustness to an inserted patch. These results demonstrate that the discovered concepts support targeted data curation to mitigate shortcut reliance (Appendix~\ref{sec:spurious_protocol}).

\subsection{Ablation Studies and Analysis}
\label{sec:ablation}

Unless otherwise specified, we conduct ablations on the Qwen3-VL-8B-Instruct model with \model for $K=10$ iterations on unlabeled iNaturalist-mini images and report top-1 accuracy with the linear probe trained on iNaturalist-mini rather than the full iNaturalist training set used in Table~\ref{tab:main_results}.

\noindent\textbf{Contrastive Reward Components.} To validate the necessity of each reward component, we ablate the positive and negative signals of the contrastive reward (Figure~\ref{fig:all_ablations}(a)). Removing the negative signal reduces the objective to measuring grounding in the source image alone, yielding generic concepts that lack distinctiveness from other instances. Eliminating the positive signal causes a sharp drop, as the optimization rewards distinctiveness indiscriminately and internalizes background noise as valid attributes. Both signals are necessary---the positive grounds concepts in the source image, while the negative promotes distinctiveness from other instances.



\noindent\textbf{Reward Modality.} To determine the optimal grounding space for the contrastive reward, we compare two modalities: \textit{Text-Text} (distance between textual concepts of different images) and \textit{Image-Text} (alignment between proposed concepts and raw visual embeddings). As shown in Figure~\ref{fig:all_ablations}(b), \textit{Image-Text} achieves substantially higher accuracy, as anchoring candidates directly against physical image features produces a cleaner preference signal than relying on intermediate linguistic representations.

\noindent\textbf{Data Augmentation.} Standard contrastive learning for visual representation learning applies augmentations~\citep{tian2020makes, simclr}. In Table~\ref{tab:augmentation}, we study which image augmentations work best for our algorithm. Surprisingly, we find that the highest accuracy is achieved with no augmentation (83.46\%), exceeding the unoptimized VLLM (69.23\%) by 14.23 percentage points. Cumulatively adding augmentations reduces accuracy to 81.92\%, while random crop alone yields the lowest accuracy (80.77\%). We therefore use original images to compute the reward. Since augmentations in contrastive learning encourage the model to be invariant to that augmentation, our result suggests that making no prior assumption about what features to be invariant is preferable for self-supervised concept discovery.


\noindent\textbf{Base Model Capacity.} To assess generalization across model scales, we evaluate the 2B, 4B, and 8B variants of Qwen3-VL-Instruct (Table~\ref{tab:base_model}). Despite similar zero-shot baselines (67\%--69\%), \model yields consistent gains of 12.7--14.2 points across all scales. Accuracy scales monotonically with model size (83.46\% for 8B), suggesting that \model benefits from stronger base models.

\noindent\textbf{Importance of Easy Negatives.} To study the effect of preference pair construction, we compare three strategies defined by the reward gap between chosen and rejected descriptions: \textit{Hard Neg.}\ (smallest gap), \textit{Random}, and \textit{Easy Neg.}\ (largest gap), each selecting 16 pairs per image (Table~\ref{tab:dpo_strategy}). Easy negatives achieve the best accuracy (83.46\%), while hard negatives degrade to 76.92\%. The self-supervised reward inherently contains noise from the continuous embedding space; forcing the model to distinguish near-identical scores amplifies this noise. Easy negatives sidestep this by isolating the widest reward discrepancies, providing an unambiguous corrective signal.

\noindent\textbf{Qualitative Analysis.}\label{sec:visualization} Figure~\ref{fig:qualitative_results} compares concepts before and after \model optimization (see Appendix~\ref{sec:more_vis} for extended visualizations). Red marks errors and visually plausible descriptions that are uninformative for the category, including generic captions and photographic filler. After \model optimization, the model suppresses these concepts and produces more specific descriptions of visible attributes in these examples.

\section{Conclusion}
We introduce \model, a framework that discovers human-readable visual concepts from unlabeled images through self-supervised preference learning. The learned concepts improve recognition across diverse domains and expose spurious cues that guide data curation. As accuracy continues to grow with more unlabeled images (Appendix~\ref{sec:dataset_size}), \model offers a scalable path toward self-supervised interpretable visual representations.

\subsubsection*{Acknowledgments}
We thank Amazon Research Award that supports this research. This work used Purdue Anvil GPU through allocation
250774 from the Advanced Cyberinfrastructure Coordination Ecosystem: Services \& Support (ACCESS) program,
which is supported by U.S. National Science Foundation
grants \#2138259, \#2138286, \#2138307, \#2137603, and
\#2138296.

\bibliographystyle{iclr2027_conference}
\bibliography{main}


\clearpage
\appendix

\section{Technical appendices and supplementary material}

\subsection{Implementation Details}
\label{sec:implementation_details}

See Table~\ref{tab:impl_details} for all implementation details.

\begin{table}[H]
\centering
\caption{\textbf{Implementation Details.}}
\label{tab:impl_details}
\small
\begin{tabular}{ll}
\toprule
\textbf{Hyperparameter} & \textbf{Value} \\
\midrule
\multicolumn{2}{l}{\textit{Concept Proposal}} \\
Base VLLM & Qwen3-VL-8B-Instruct \\
Sampling temperature & 2.0 \\
Top-$p$ / Top-$k$ & 1.0 / 100 \\
Repetition penalty & 1.1 \\
Candidate descriptions per image & 16 \\
\midrule
\multicolumn{2}{l}{\textit{Contrastive Reward}} \\
Cross-modal encoder & OpenCLIP ViT-H/14 (LAION-2B) \\
Reward batch size & 1,300 \\
Reward temperature $\tau$ & 0.07 \\
\midrule
\multicolumn{2}{l}{\textit{Preference Pairing}} \\
Pairing strategy & Easy Neg. \\
Candidate pairs per image & $\binom{16}{2} = 120$ \\
Selected pairs per image & 16 \\
\midrule
\multicolumn{2}{l}{\textit{DPO Training}} \\
DPO $\beta$ & 0.05 \\
DPO loss & Sigmoid \\
Epochs per iteration & 3 \\
Global batch size & 512 \\
LLM / Merger learning rate & $1 \times 10^{-6}$ \\
Vision tower learning rate & $5 \times 10^{-6}$ \\
LR schedule & Constant with 1\% warmup \\
Weight decay & 0.01 \\
Precision & bf16 \\
\midrule
\multicolumn{2}{l}{\textit{Pipeline}} \\
Total iterations & 10 \\
GPUs & $8 \times$ RTX 6000 Blackwell Pro \\
Total training time & 22 hours (176 GPU hours) \\
\midrule
\multicolumn{2}{l}{\textit{Evaluation}} \\
Sampling temperature & 0.0 \\
Repetition penalty & 1.0 \\
Deduplication threshold & 0.65 \\
Classifier & Logistic Regression (LBFGS) \\
Max iterations & 5,000 \\
L2 regularization $\lambda$ & $1 \times 10^{-3}$ \\
\bottomrule
\end{tabular}
\end{table}

\subsection{Algorithm}
\label{sec:algorithm}

The complete optimization procedure described in Section~\ref{sec:method} is detailed in Algorithm~\ref{alg:ssdpo}.

\begin{algorithm}[H]
\caption{\model: Self-Supervised Concept Discovery via Preference Learning}\label{alg:ssdpo}

\DontPrintSemicolon

\KwIn{Unlabeled dataset $\mathcal{X}$, Initial VLLM concept generator $\pi_0$, Frozen cross-modal encoder $\mathcal{E}$ (e.g., CLIP), Iterations $K$, Batch size $B$, Candidate descriptions per image $N$}
\KwOut{Optimized concept generator $\pi_K$}

\tcp{Iterative self-supervised discovery loop}
\For{$k = 0, \ldots, K-1$}{
    $\mathcal{D}_k \gets \emptyset$ \tcp*{Initialize preference dataset for iteration $k$}

    Sample a batch of unlabeled images $\mathcal{B} = \{x_1, \ldots, x_B\} \sim \mathcal{X}$ \;

    \tcp{Encode original images with the frozen vision encoder}
    $v_x \gets \mathcal{E}(x)$ for each $x \in \mathcal{B}$; $\mathcal{V} \gets \{v_x : x \in \mathcal{B}\}$ \;

    \For{\textbf{each} anchor image $x \in \mathcal{B}$}{
        \tcp{1. Sample candidate descriptions}
        Sample $\mathcal{C}(x) = \{c_1, \ldots, c_N\} \sim \pi_k(\cdot \mid x)$ using high-temperature decoding \;

        \tcp{2. Cross-Modal Contrastive Scoring}
        \For{\textbf{each} description $c \in \mathcal{C}(x)$}{
            $t_c \gets \mathcal{E}(c)$ \tcp*{Extract normalized text embedding}

            \tcp{Score grounding and distinctiveness}
            $R(c, x) \gets \frac{t_c^\top v_x}{\tau} - \log \sum_{v \in \mathcal{V}} \exp \left( \frac{t_c^\top v}{\tau} \right)$ \;
        }

        \tcp{3. Easy Negative Preference Selection}
        $\mathcal{P}_x \gets$ the 16 pairs $(c_i,c_j)$ in $\mathcal{C}(x)$ with the largest absolute reward gaps $|R(c_i,x)-R(c_j,x)|$ \;

        \For{\textbf{each} $(c_i,c_j) \in \mathcal{P}_x$}{
            $c_w \gets \arg\max_{c \in \{c_i, c_j\}} R(c, x)$ \tcp*{Higher-scoring description}
            $c_l \gets \arg\min_{c \in \{c_i, c_j\}} R(c, x)$ \tcp*{Lower-scoring description}

            $\mathcal{D}_k \gets \mathcal{D}_k \cup \{(x, c_w, c_l)\}$ \;
        }
    }

    \tcp{4. Preference Optimization for Autonomous Discovery}
    Update $\pi_{k+1}$ by minimizing the DPO loss $\mathcal{L}_{\text{DPO}}(\pi_{\theta}; \pi_k)$ on dataset $\mathcal{D}_k$ \;
}

\Return $\pi_K$ \;

\end{algorithm}

\subsection{Comparison with Latent Self-Supervised Learning}
\label{sec:simclr_compare}

\begin{table}[H]
    \centering
    \caption{\textbf{Comparison with Latent Self-Supervised Learning.} Linear probing accuracy comparing the continuous, black-box representations of SimCLR against our discrete, interpretable concepts. Our approach yields a 20.4-point absolute improvement over the fine-tuned latent baseline.}
    \label{tab:simclr_compare}
    \setlength{\tabcolsep}{4mm}
    \begin{tabular}{lc}
        \toprule
        \textbf{Method} & \textbf{Top-1 Accuracy (\%)} \\
        \midrule
        SimCLR (Pretrained ResNet-50) & 55.00 \\
        SimCLR (Fine-tuned ResNet-50) & 63.08 \\
        \textbf{\model{} (Ours)} & \textbf{83.46} \\
        \bottomrule
    \end{tabular}
\end{table}

To benchmark our interpretable visual lexicons against standard latent self-supervised learning, we compare our framework with SimCLR~\citep{simclr}. We first evaluate a pretrained SimCLR ResNet-50 backbone by removing its projection head and training a linear probe directly on the frozen representations. We then fine-tune this backbone on our unlabeled dataset by initializing a new projection head before applying the identical linear probing protocol. Both SimCLR baselines strictly follow the data augmentation pipeline detailed in the original implementation.

Table~\ref{tab:simclr_compare} shows that our method outperforms both SimCLR baselines. Fine-tuning improves SimCLR accuracy from 55.00\% to 63.08\%, while our approach achieves 83.46\%, a gain of 20.4 percentage points over the fine-tuned baseline. These results show that \model learns interpretable concepts with strong discriminative performance relative to the evaluated SimCLR representations. Since the methods differ in architecture and pretraining, this comparison does not isolate the effect of the concept bottleneck.

\subsection{Cross-Modal Encoder Capacity}
\label{sec:clip_encoder}

To evaluate sensitivity to the cross-modal encoder, we ablate the frozen CLIP architecture used for the contrastive reward (Table~\ref{tab:clip_encoder}). Both encoders yield strong gains over the unoptimized VLLM, with ViT-H/14 achieving a modest advantage over ViT-B/16 (83.46\% vs.\ 82.31\%).

\begin{table}[H]
\centering
\caption{\textbf{Impact of Cross-Modal Encoder Capacity.} Ablation over the frozen CLIP encoder used for the contrastive reward.}
\label{tab:clip_encoder}
\small
\begin{tabular}{lccc}
\toprule
\multirow{2}{*}{\textbf{Encoder}} & \multirow{2}{*}{\textbf{Unoptimized VLLM}} & \multicolumn{2}{c}{\textbf{VLLM + \model}} \\
\cmidrule(lr){3-4}
& & \textbf{ViT-B/16} & \textbf{ViT-H/14} \\
\midrule
\textbf{Top-1 Accuracy (\%)} & 69.23 & 82.31 & \textbf{83.46} \\
\bottomrule
\end{tabular}
\end{table}

\subsection{Scaling Unlabeled Data}
\label{sec:dataset_size}

We vary the number of unlabeled training images from 100 to 1,300 (Table~\ref{tab:dataset_size}). Accuracy rises from 76.54\% to 83.46\% and peaks at the largest scale, suggesting that \model will continue to benefit from larger unlabeled collections.

\begin{table}[H]
\centering
\caption{\textbf{Impact of Unlabeled Dataset Size.} Accuracy has not saturated at 1,300 unlabeled images, suggesting that \model scales effectively with more unlabeled data.}
\label{tab:dataset_size}
\small
\begin{tabular}{lcccccc}
\toprule
\multirow{2}{*}{\textbf{Dataset Size}} & \multirow{2}{*}{\textbf{Unoptimized}} & \multicolumn{5}{c}{\textbf{+ \model}} \\
\cmidrule(lr){3-7}
& & 100 & 400 & 700 & 1000 & 1300 \\
\midrule
Top-1 Acc. (\%) & 69.23 & 76.54 & 80.38 & 81.15 & 81.15 & \textbf{83.46} \\
\bottomrule
\end{tabular}
\end{table}

\subsection{ISIC Shortcut-Mitigation Protocol}
\label{sec:spurious_protocol}

The concept generator remains frozen throughout the ISIC experiment. For downstream filtering, we rank the discovered concept clusters by training-set AUROC against SONIC group membership (SONIC is the ISIC Archive collection of benign lesions photographed with colored patches) and retain the top ten. Each training image receives the maximum activation over these clusters. We set the threshold to match the fraction of training descriptions that mention a patch, remove the 6,463 flagged images, and retrain the linear probe using the same concept bank and probe hyperparameters. Group membership is used for cluster selection and detection evaluation in this application, but not for training the concept generator or generating concepts. The error rate under patch insertion is computed separately for each probe over the malignant test images that it classifies correctly before patch insertion.

\subsection{VLLM-as-a-Judge Protocol}
\label{sec:judge_protocol}

We evaluate a fixed set of 39 before-and-after examples spanning natural images, galaxies, and blood-cell microscopy. This set is not a random sample of the eight classification benchmarks. GPT-5.6-Sol (model identifier \texttt{gpt-5.6-sol}, accessed through the OpenAI Codex CLI) receives each image and its two concept lists without model identities or class labels. The prompt asks which list describes the main subject more accurately and specifically, penalizing incorrect, unsupported, generic, or repetitive details. It instructs the judge to disregard background, lighting, camera style, and image quality, and not to favor longer descriptions.

Each pair is evaluated in both A/B orders with medium reasoning effort, and the judge is constrained to return a JSON verdict of A or B. Win rates in Table~\ref{tab:vlm_judge} are computed over the resulting 78 judgments (39 images $\times$ 2 orders); e.g., the evaluator prefers \model over the unoptimized VLLM in 67 of them (85.9\%). The result measures preference on this example set, not a benchmark-wide quality estimate.

\paragraph{Baseline concept lists.} The four baselines do not describe individual images; each constructs a dataset-level concept bank and grounds it with CLIP. To obtain a per-image concept list for a baseline, we take the concept bank it produced for the dataset the image belongs to (iNaturalist, CUB, Galaxy10, or BloodMNIST) and select the $k$ bank entries with the highest OpenCLIP ViT-H/14 image-text cosine similarity, where $k$ equals the number of phrases in the \model description for that image so that list lengths match. LaBo uses its submodularly selected concepts, LF-CBM its filtered concept set, DCLIP the union of its per-class descriptors, and TextSpan the descriptions selected by its spanning procedure (the full general description set for CUB and Galaxy10, whose spanning sets were not retained). These lists are then judged against the \model concepts under the same protocol. Note that LF-CBM and DCLIP concept banks are conditioned on class names, so their concepts occasionally contain taxonomic identifiers (e.g., ``Arctostaphylos''), which the judge credits when correct.

\subsection{Quantitative Evolution of Concept Diversity}
\label{sec:concept_stats}

\begin{figure}[H]
    \centering
    \includegraphics[width=1.0\textwidth]{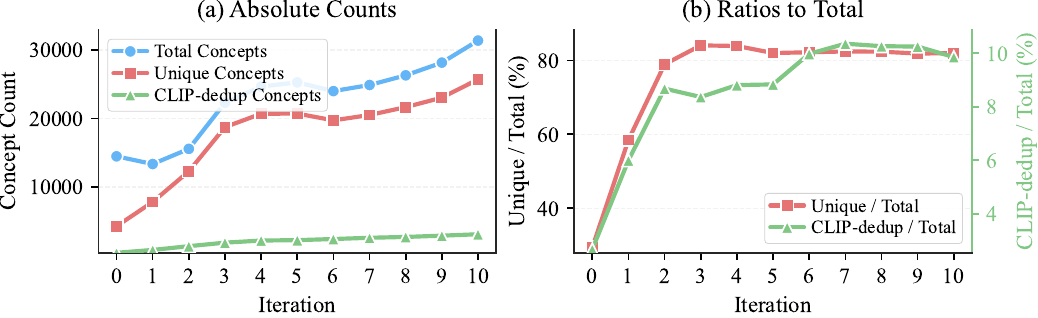}
    \caption{\textbf{Evolution of Concept Diversity.} We track the absolute counts (a) and ratios (b) of unique and CLIP-deduplicated concepts generated across optimization iterations. Our optimized concept generator progressively escapes generic mode collapse, expanding its vocabulary to produce a richer, highly distinct, and less redundant visual lexicon.}
    \label{fig:concept_stats}
\end{figure}

To demonstrate that \model successfully suppresses repetitive conversational priors, we track the diversity of the generated concept pool across iterations. Figure~\ref{fig:concept_stats} illustrates the absolute counts and ratios of unique and CLIP-deduplicated concepts. The unoptimized base model (Iteration 0) suffers from generic mode collapse, yielding a highly redundant lexicon with a low uniqueness ratio. As optimization progresses, both the absolute volume and the proportion of distinct concepts increase substantially and stabilize. This suggests that the contrastive reward encourages the VLLM to actively explore the long tail of its vocabulary, expanding its capacity to articulate granular visual details.

\subsection{Qualitative Trajectory of Concept Refinement}
\label{sec:concept_trajectory}

\begin{figure}[t!]
    \centering
    \includegraphics[width=1.0\textwidth, trim={0.7cm 0 1.6cm 0}, clip]{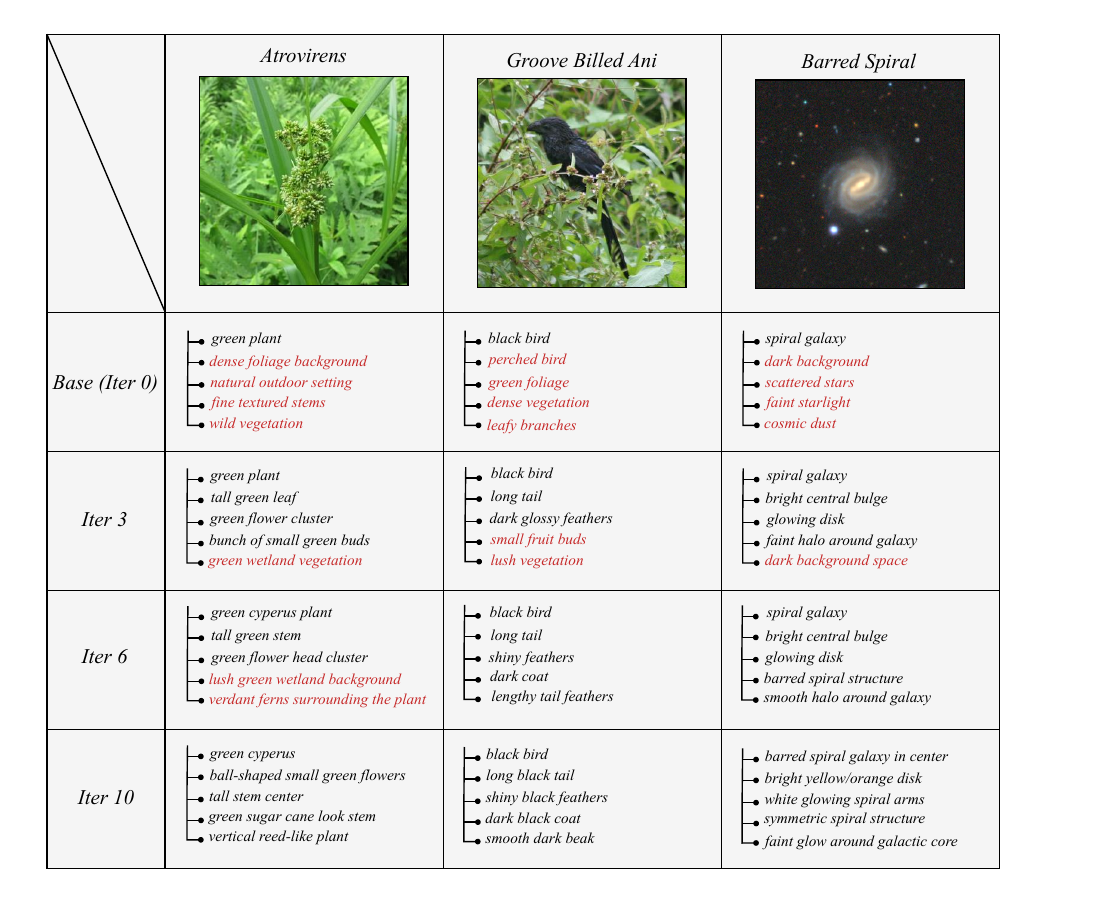}
    \caption{\textbf{Trajectory of Concept Refinement.} For each sample, we trace the evolution of generated concepts from the unoptimized base model (Iter 0) through successive \model iterations. \textcolor{red}{Red} marks concepts that are incorrect or uninformative for the category. Our optimized model suppresses these concepts and produces more specific visual attributes.}
    \label{fig:concept_changes}
\end{figure}

Figure~\ref{fig:concept_changes} qualitatively tracks how hypothesized concepts for individual images evolve throughout the \model loop. Initially, the unoptimized base model defaults to broad, non-discriminative linguistic priors. Through successive iterations, the optimization systematically discards these generic labels, refining the lexicon into increasingly precise descriptions of visible structures.

\subsection{Extended Visualizations Across Domains}
\label{sec:more_vis}

\begin{figure}[t!]
    \centering
    \includegraphics[width=\textwidth]{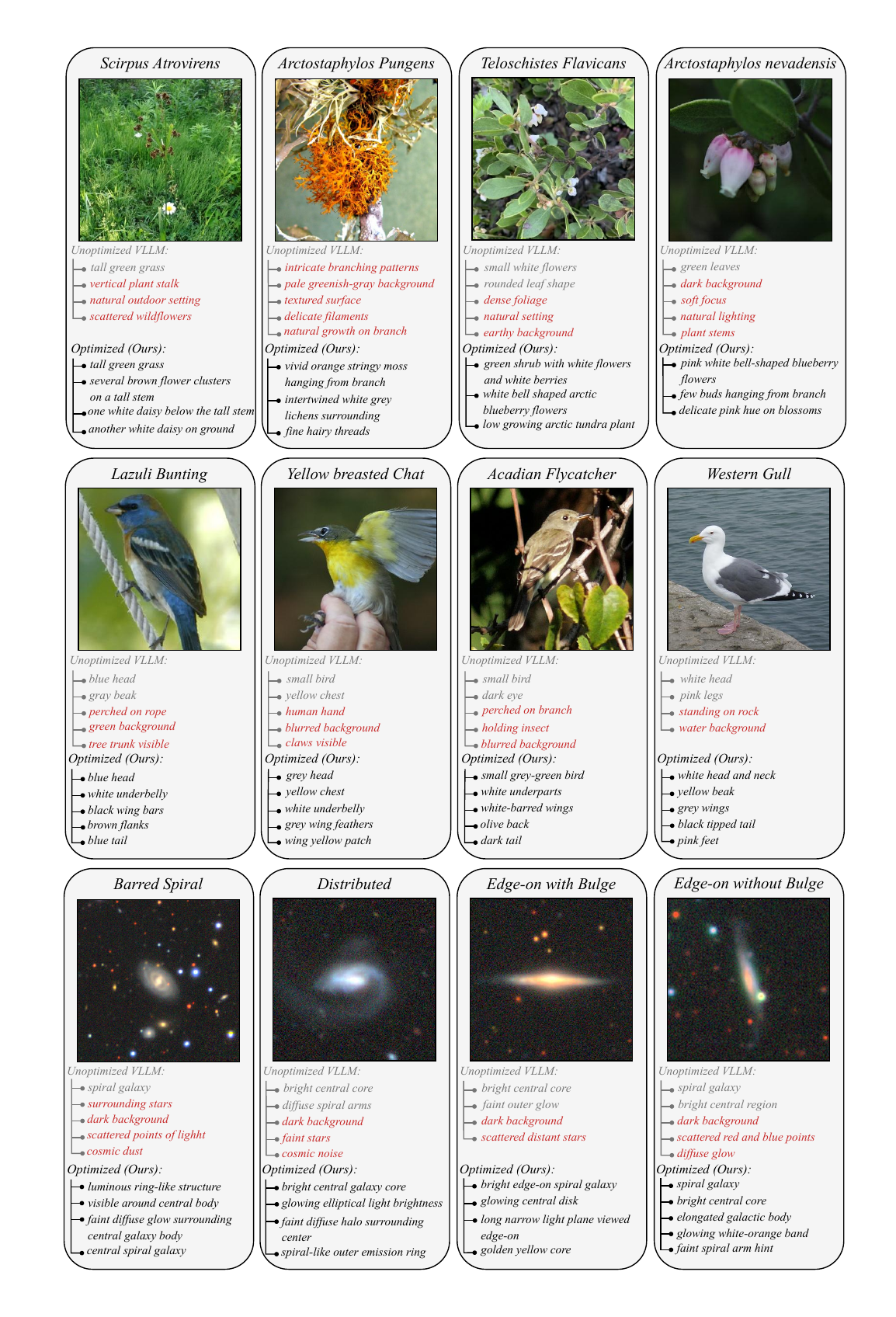}
    \vspace{-0.8cm}
    \caption{\textbf{Visualizing Self-Supervised Concept Discovery.} For each sample, we contrast the top concepts generated by the unoptimized base model (top list) with our optimized concept generator (bottom list). \textcolor{red}{Red} marks concepts that are incorrect or uninformative for the category. Our optimized model suppresses these concepts and produces more specific visual attributes.}
    \label{fig:extended_vis}
\end{figure}

Figure~\ref{fig:extended_vis} compares concepts from the unoptimized VLLM and the optimized concept generator across our evaluation domains. The examples show a shift from generic descriptions toward more specific visual attributes.

\subsection{Human Study Samples}
\label{sec:human_study_samples}

\begin{figure}[h]
\centering
\includegraphics[width=\textwidth]{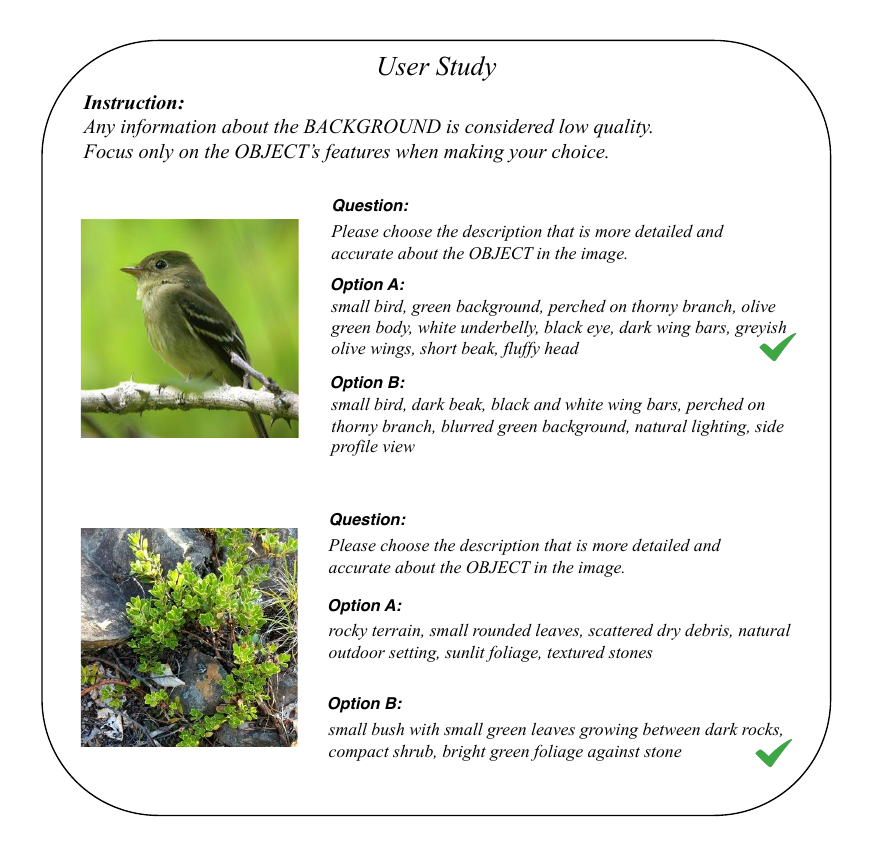}
\caption{\textbf{User Study Samples.} Two sample questions from our user study, which are also given to the VLLM judge. Participants follow the instruction and select among two anonymized concept lists.}
\label{fig:human_study_samples}
\end{figure}

Figure~\ref{fig:human_study_samples} shows two sample questions, which both the human evaluators and the VLLM judge (Appendix~\ref{sec:judge_protocol}) answer.

\subsection{Limitations}
\label{sec:limitations}

Our concept discovery quality depends on the capacity of the frozen CLIP encoder and the VLLM's pre-trained vocabulary, both of which can be improved by adopting stronger foundation models as they become available. We train on a single source domain (iNaturalist) and observe strong cross-domain transfer; exploring diverse source domains is a natural next step. As shown in Table~\ref{tab:dataset_size}, accuracy improves overall with the number of unlabeled training images and has not yet saturated at 1,300 images, suggesting that further gains are achievable with larger-scale unlabeled data.

\subsection{Broader Impacts}
\label{sec:broader_impacts}

Our work advances interpretable AI by enabling concept discovery without human annotation, which has positive implications for high-stakes domains such as medical imaging and scientific discovery, where transparent reasoning is essential. By removing the reliance on labeled data, our framework also lowers the barrier to deploying interpretable models in data-rich but annotation-scarce settings. On the other hand, the discovered concepts inherit the biases present in the VLLM's pre-training data; practitioners should be aware of this when applying the framework to sensitive domains.


\clearpage

\end{document}